\documentclass[]{interact}
\usepackage[utf8]{inputenc}

\usepackage{epstopdf}
\usepackage[caption=false]{subfig}

\usepackage[longnamesfirst,sort]{natbib}
\bibpunct[, ]{(}{)}{;}{a}{,}{,}
\renewcommand\bibfont{\fontsize{10}{12}\selectfont}

\usepackage{amsmath,amssymb}

\newcommand{\targetuniversitymention}{National Tsing Hua University (NTHU)}
\newcommand{\targetteamphrase}{the National Tsing Hua University (NTHU) football team}
\newcommand{\targetteamshort}{NTHU}
\newcommand{\targetdomainshort}{NTHU}
\newcommand{\Targetdomainshort}{NTHU}
\newcommand{\targetdomainlong}{NTHU university football}
\newcommand{\targetuniversitydata}{NTHU university data}
\newcommand{\targetuniversitydataset}{NTHU university dataset}
\newcommand{\targetdataset}{NTHU dataset}
\newcommand{\targeteventfeatures}{NTHU event features}

\begin{document}

\articletype{ARTICLE}

\title{Interpretable Machine Learning for Football Performance Analysis:
Evidence of Limited Transferability from Elite Leagues to University Competition}

\author{
\name{Yu-Fang Tsai\textsuperscript{a}*, 
Yu-Jen Chen\textsuperscript{b}*\thanks{*These authors contributed equally to this paper.\\ CONTACT Yu-Jen Chen, email: \texttt{yujenchen@gapp.nthu.edu.tw}}, 
Kok-Hua Tan\textsuperscript{c}, 
Sheng-Chieh Huang\textsuperscript{d}, 
You-Ying Ji\textsuperscript{e}, 
Yu-Lun Chen\textsuperscript{f}, 
Chun-Yi Wang\textsuperscript{g}
and Chien-Ming Hsu\textsuperscript{a}}
\affil{\textsuperscript{a} Department of Kinesiology, National Tsing Hua University, Hsinchu, Taiwan;\\
\textsuperscript{b} Department of Computer Science, National Tsing Hua University, Hsinchu, Taiwan;\\
\textsuperscript{c} Physical Education Office, National Tsing Hua University, Hsinchu, Taiwan;\\
\textsuperscript{d} Interdisciplinary Program of Management and Technology, National Tsing Hua University, Hsinchu, Taiwan;\\
\textsuperscript{e} Department of Physics, National Tsing Hua University, Hsinchu, Taiwan;\\
\textsuperscript{f} Interdisciplinary Program of Electrical Engineering and Computer Science, National Tsing Hua University, Hsinchu, Taiwan;\\
\textsuperscript{g} Department of Mathematics, National Tsing Hua University, Hsinchu, Taiwan
}
}

\maketitle
\begin{abstract}
Machine learning has become increasingly prevalent in football performance analysis, yet most studies prioritize predictive accuracy while implicitly assuming that learned performance determinants and their interpretations are transferable across competition levels. Whether interpretability remains reliable under domain shift—from elite to university football remains largely unexplored.
This study investigates whether performance determinants learned from elite competitions are structurally transferable to university-level football and whether their interpretations remain robust under domain shift. Models were trained on large-scale event data from the top five European leagues and applied to university football data from \targetuniversitymention{} using an identical feature space. Random Forest and Multilayer Perceptron models were interpreted using SHapley Additive exPlanations (SHAP) and Counterfactual Impact Score (CIS).
Across five experiments, elite football exhibited a stable and consistent hierarchy of performance determinants across leagues, models, and explanation methods. In contrast, \targetdomainlong{} showed substantial reordering of key indicators, reduced explanation stability, weaker structural agreement with elite domains, and increased sensitivity to explanation method.
These findings suggest that interpretability robustness is domain-dependent. Rather than reflecting methodological limitations alone, instability in explanations under domain shift may serve as a diagnostic signal of structural ambiguity in the target domain.
\end{abstract}


\section{Introduction}

The analysis of football performance has evolved substantially over the past decades, progressing from early notational studies to advanced data-driven and machine learning–based methodologies. Foundational work demonstrated that match outcomes arise from a complex interplay between skill and stochastic factors \citep{ReepBenjamin1968}, motivating the systematic collection and analysis of match events. Subsequent developments formalized match analysis as a structured discipline aimed at identifying performance indicators associated with success \citep{Carling2007}.

With the increasing availability of large-scale event data, football performance analysis has increasingly adopted machine learning techniques to model match outcomes, team effectiveness, and tactical behavior. Systematic reviews indicate that elite football competitions exhibit relatively stable performance determinants—such as possession control, shooting efficiency, and defensive organization—that consistently appear across leagues and seasons \citep{Sarmento2014,Sarmento2018}. These findings have reinforced the perception that data-driven models can uncover meaningful and generalizable indicators of football performance.

However, most existing studies implicitly assume that performance indicators learned from elite competitions are transferable across different competitive contexts. This assumption is rarely examined explicitly, despite substantial differences in tactical strategy, organizational stability, and player experience across competitions. In particular, university-level football operates under markedly different conditions, including limited training time, heterogeneous player skill levels, and variable team composition. These differences raise fundamental questions regarding whether performance structures observed in elite football remain valid at lower competitive levels.

Recent work has emphasized that football performance should be understood not only in terms of predictive accuracy but also through interpretability and contextual meaning \citep{ReinMemmert2016,GomezRuano2020}. As machine learning models grow more complex, explainable artificial intelligence (XAI) techniques are increasingly used to attribute predictions to specific performance indicators. These explanations are often treated as transparent representations of underlying performance determinants and are frequently used to inform practical decision-making.

However, interpretability is not just a built-in property of a model; it depends on how the model, the explanation method, and the data domain interact. Under domain shift, where the deployment data distribution differs from the training data, feature importance explanations may become unstable or inconsistent. While such instability is often seen as methodological noise or model unreliability, another perspective is that it may reflect real uncertainty in the target domain, especially in competitions where tactical and performance hierarchies are not yet well established.

From this perspective, interpretability itself becomes an analyzing method rather than a guaranteed property. Understanding when explanations are stable, transferable, and method-invariant is essential before translating model outputs into actionable insights. This issue is especially critical in university football, where performance structures may still be emerging.

Accordingly, this study does not aim to develop a superior match outcome prediction model. Instead, it investigates whether football performance determinants learned from elite competitions are structurally transferable to university-level football, and whether interpretability remains reliable under such domain shift. To isolate domain shift effects, models are trained exclusively on elite football data and applied to university football only for inference. This design enables a controlled comparison of performance structures across competition levels.

This study makes three contributions. First, it provides empirical evidence that performance determinants exhibit a stable and shared structure across elite competitions but diverge substantially at the university level. Second, it demonstrates that explanation stability and agreement are domain-dependent, with interpretability being robust in elite football but less reliable under domain shift. Third, it reframes explanation instability as an indicator of structural uncertainty, highlighting the need to evaluate interpretability before applying model insights across competition levels.

\section{Related Work}
\subsection{Performance Analysis in Football}

Football performance analysis began with notational and statistical studies that examined the role of skill and chance in match outcomes \citep{ReepBenjamin1968}. Later work established match analysis as a systematic process that combines structured observation, event coding, and contextual interpretation \citep{Carling2007}. Methodological guidance was further developed in applied texts and critical reviews \citep{ODonoghue2009,MackenzieCushion2013}. Together, these studies built the foundation for identifying match actions associated with success.

As large-scale event data became available, the field moved toward data-driven and machine learning approaches \citep{Moya2025,Wong2025,Hewitt2023}. Reviews of elite football have reported recurring determinants of performance, such as possession control, shooting efficiency, and defensive organization \citep{Sarmento2014,Sarmento2018}. At the same time, scholars have argued that analysis should go beyond outcome prediction and account for tactical and organizational context \citep{ReinMemmert2016,GomezRuano2020,Biermann2025}. This raises an important question: do these structures remain stable across competition levels?

\subsection{Data-Driven and Machine Learning Approaches}

Many studies use machine learning to predict match outcomes and evaluate team performance in professional football. Methods range from conventional models to deep learning and are typically built on event-based features \citep{PantzalisTjortjis2020,Li2020,AttaMills2024}. Related work has also evaluated player contributions at the action level and produced data-driven player rankings \citep{Decroos2019,Pappalardo2019}.

Recent research has expanded to player behavior and team interaction dynamics, emphasizing the temporal nature of football performance \citep{Kusmakar2020,GarciaAliaga2021,Yeung2025}. Hybrid data- and knowledge-driven frameworks have also been proposed \citep{Berrar2024}. However, most studies validate models within the same domain used for training, so cross-level transferability remains underexplored.

\subsection{Explainable Artificial Intelligence in Football Analytics}

As models become more complex, explainable artificial intelligence (XAI) has become increasingly important for football analytics \citep{Molnar2022}. Methods such as SHAP \citep{LundbergLee2017} and DeepLIFT \citep{Shrikumar2017} attribute predictions to input features and improve model transparency. In football, these methods have been used to identify key indicators and support tactical interpretation \citep{Javed2023,Moustakidis2023,Cavus2023,Ma2025}.

Beyond model-level applications, conceptual and systems-oriented studies discuss how XAI can support decision making in practice \citep{Procopiou2023,Wang2024}. Methodological work has also proposed alternative explanation pipelines, including knockoff-based approaches for better robustness under feature dependence \citep{Zhao2025}.

Despite this progress, explanation outputs are often treated as inherently stable and transferable. Differences in explanations are commonly framed as model noise rather than potential evidence of structural data differences. As a result, the robustness of interpretability under domain shift remains insufficiently tested in football research.

\subsection{Limitations of Transferability Across Competition Levels}

Competitive level is known to affect physical, technical, and tactical characteristics of football performance \citep{Franca2022,Stafylidis2024}. This suggests that determinants learned from elite football may not directly generalize to university or amateur contexts.

Nevertheless, most data-driven football studies do not explicitly test whether elite-derived determinants remain valid after domain shift. This is a key limitation, given the growing use of elite-trained analytical frameworks in non-elite settings.



\section{Methods}
\subsection{Overall Framework and Study Design}

This study examines the structural transferability of football performance determinants and the robustness of interpretability under domain shift. An elite-to-university transfer framework is adopted, in which models are trained exclusively on elite football data and applied to university football for inference and comparative analysis.

\begin{figure}[t]
    \centering
    \includegraphics[width=0.95\linewidth]{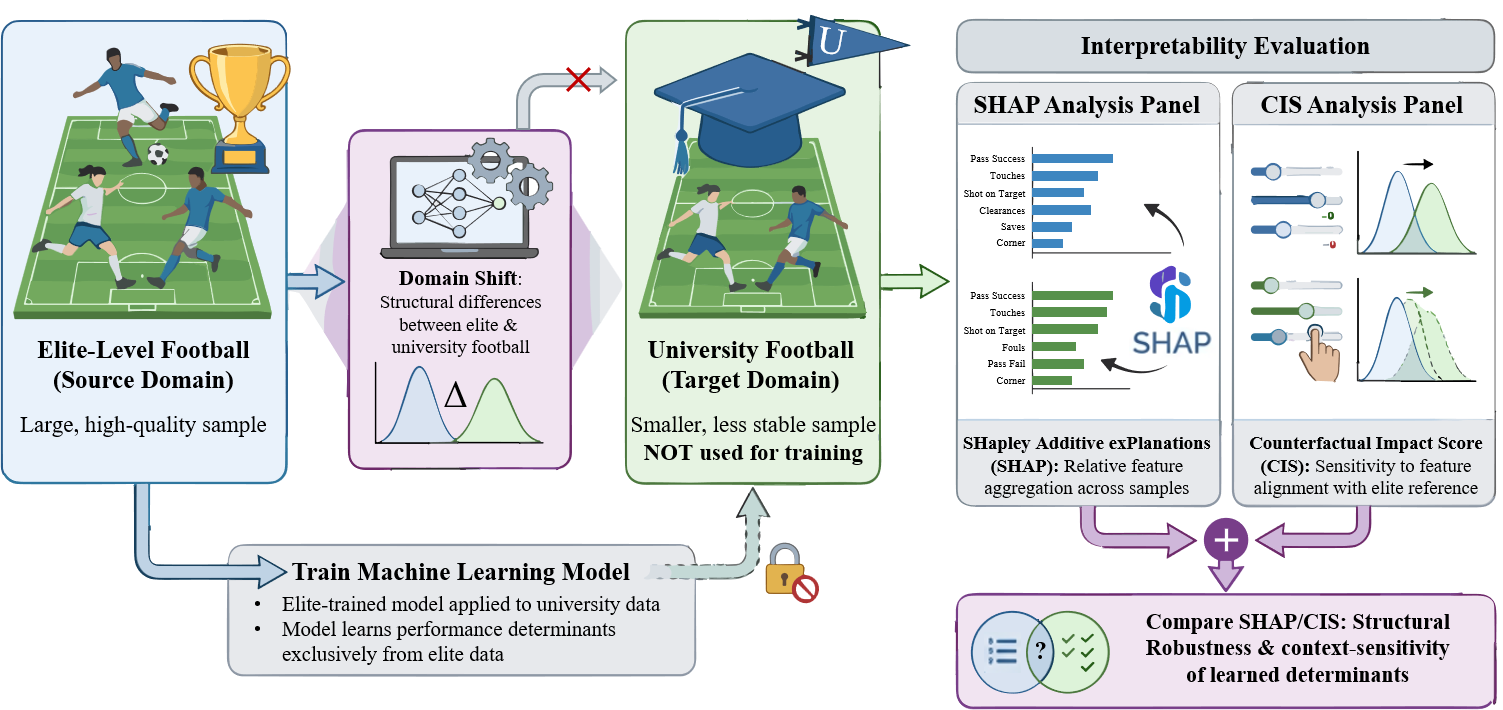}
    \caption{Overall framework and study design. Models are trained exclusively on elite football data and applied to university football for inference only. This design isolates domain shift effects and enables evaluation of whether learned performance structures and their explanations remain stable across competition levels.}
    \label{fig:overall_framework}
\end{figure}

In this framework, elite football is treated as the source domain for learning performance structures, while university football serves as the target domain for interpretability evaluation. This design enables direct comparison of explanation structures across domains. The overall framework is illustrated in Figure~\ref{fig:overall_framework}.

\subsection{Data Sources and Domains}

\subsubsection{Elite Football Data}

Elite football data were collected from the top five European leagues: Premier League, La Liga, Serie A, Bundesliga, and Ligue 1. The dataset covers three recent seasons (2022/23, 2023/24, and 2024/25), comprising a large number of matches with consistent event annotations.

Match event data were retrieved from public WhoScored match pages through the \texttt{soccerdata} Python package. The study uses this public-web access channel only; no direct licensed Opta/Stats Perform feed was queried for model training. Event semantics were aligned to Opta-style definitions.

For provenance control, elite data were collected during a single scripted extraction window from June to July 2025. Exclusion rules were predefined: fixtures were removed if (i) the match did not have a finalized scoreline at extraction, (ii) event tables could not be parsed after repeated retries, or (iii) duplicate fixture IDs were returned across repeated pulls. Excluded fixtures were not imputed.

All elite data were processed into match-level event counts using consistent feature definitions across leagues and seasons, and to avoid information leakage between team perspectives, data splitting was performed match-wise before perspective expansion. Elite matches were randomly split into training, validation, and test sets (8:1:1) at the match level, and each split was then expanded into two samples per match (home and away perspectives), resulting in doubled sample counts within each split. This dataset was used exclusively as the source domain for model development and held-out source-domain evaluation.

The number of matches collected for each league and season is shown in Table~\ref{tab:elite_provenance}.

\begin{table}[h]
\tbl{Elite-dataset provenance and completeness summary (WhoScored via \texttt{soccerdata} snapshot).}
{\begin{tabular}{p{0.15\linewidth}ccccp{0.35\linewidth}}
\toprule
League & 2022/23 & 2023/24 & 2024/25 & Total & Completeness note \\
\midrule
Premier League & 380 & 380 & 380 & 1140 & Complete in this snapshot. \\
La Liga & 380 & 380 & 380 & 1140 & Complete in this snapshot. \\
Ligue 1 & 379 & 380 & 306* & 1065 & One match unavailable in 2022/23. \\
Bundesliga & 304 & 306 & 306 & 916 & Two matches unavailable in 2022/23. \\
Serie A & 375 & 380 & 380 & 1135 & Five matches unavailable in 2022/23. \\
\bottomrule
\end{tabular}}
\begin{minipage}{0.95\linewidth}
\footnotesize\emph{Note:} * In the 2024/25 season, Ligue 1 reduced from 20 to 18 teams; therefore, the full-season match count decreased from 380 to 306.
\end{minipage}
\label{tab:elite_provenance}
\end{table}

\subsubsection{University Football Data}

University-level data were obtained from 17 official matches played by \targetteamphrase{} over a three-year period. At the time of this study, university football competition in Taiwan was organized into four competitive levels, with \targetteamshort{} competing in the second division. The authors confirmed that all 17 matches included in this study were drawn exclusively from this division, thereby ensuring competitive consistency within the target domain. Although these matches spanned diverse competition formats, playing surfaces, and match durations, they collectively provide representative coverage of university-level playing conditions.

Event annotations were manually performed by three certified referees following OPTA guidelines. To ensure reliability and reduce subjective bias, final event counts were determined by majority agreement across the three referees. In cases where all three counts differed, the median value was used as the adjudicated count.

The university dataset was used exclusively for inference and interpretability analysis. Although limited in size, it serves as a target domain for structural evaluation rather than statistical generalization. 

\begin{figure}[h]
    \centering
    \includegraphics[width=0.95\linewidth]{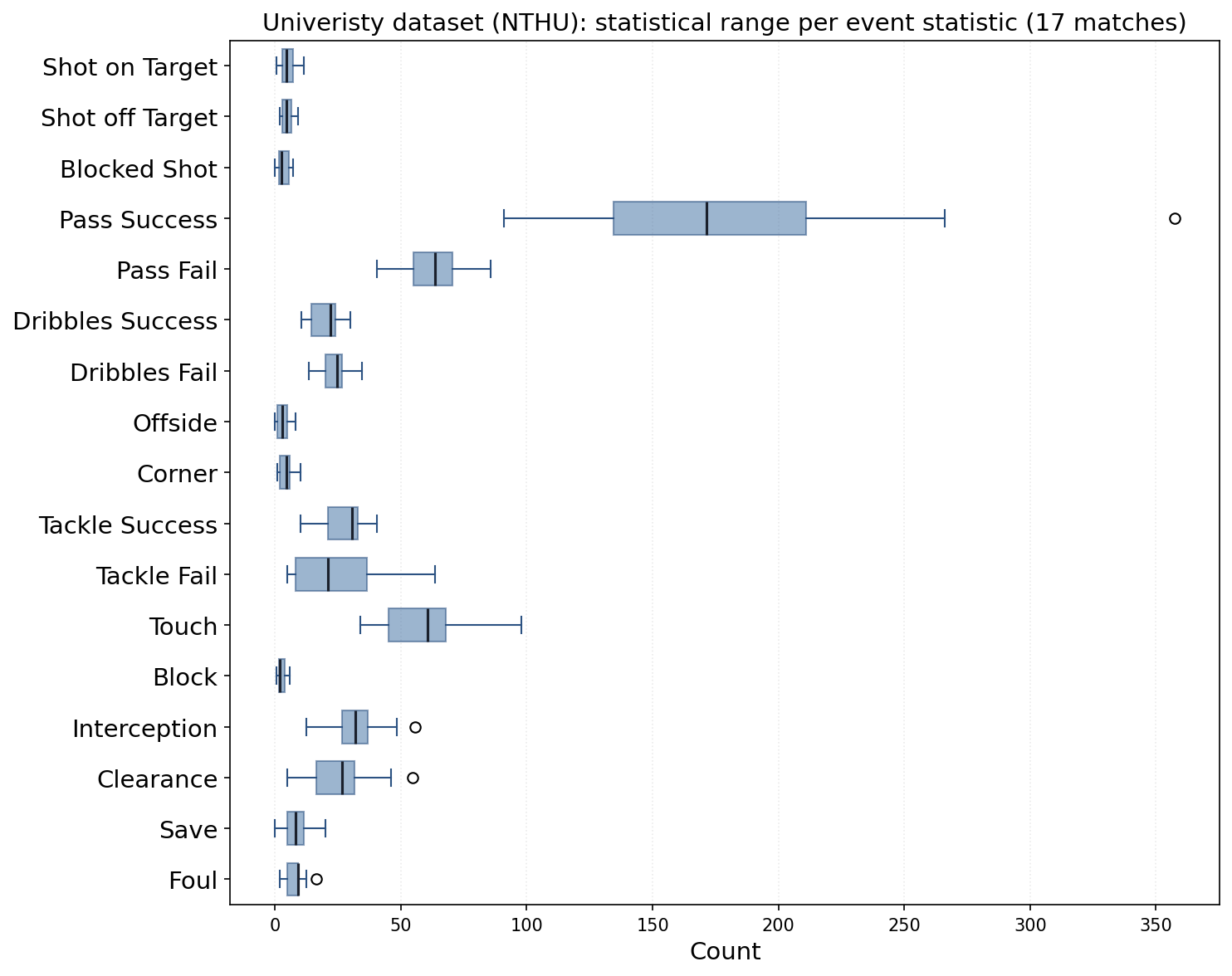}
    \caption{Statistical range of \targetdomainlong{} event features across the 17 matches. The figure provides a compact view of feature-level dispersion in the target domain.}
    \label{fig:nthu_statistical_range}
\end{figure}

To characterize target-domain variability, Figure~\ref{fig:nthu_statistical_range} summarizes the statistical range of \targeteventfeatures{} across matches. From the figure, passing-related and defensive events show substantial variability across matches, indicating less consistent match dynamics at the university level. Overall, the target domain exhibits greater heterogeneity, which may contribute to instability in learned patterns and model explanations under domain shift.

\subsection{Feature Representation}

A total of 19 event categories were recorded for each match. Among these, \textit{Goal} and \textit{Enemy Goal} were used to define the prediction target as goal difference:

\begin{equation}
y = \text{Goal} - \text{Enemy Goal}
\end{equation}

This formulation yields a single-output regression task, where positive values indicate winning performance and negative values indicate losing performance. The remaining 17 event categories were used as input features for model training.

For completeness, the full set of recorded events is listed below:

\textbf{Offensive events:}
Shot on Target, Shot off Target, Blocked Shot, Pass Success, Pass Fail, Dribbles Success, Dribbles Fail, Offside, Corner, and Goal.

\textbf{Defensive events:}
Tackle Success, Tackle Fail, Touch, Block, Interception, Clearance, Save, Foul, and Enemy Goal.

Event counts were aggregated at the match level to form fixed-length feature vectors. Consistent feature definitions were applied across elite and university datasets to ensure comparability of model behavior and explanation outputs. Feature selection prioritized interpretability and domain consistency, enabling direct comparison of learned performance structures.

\subsection{Machine Learning Models}

Multiple machine learning models were evaluated, including both tree-based and neural network architectures:

\begin{itemize}
  \item Decision Tree (DT) \citep{Breiman1984}
  \item Multilayer Perceptron (MLP) \citep{Murtagh1991}
  \item Random Forest (RF) \citep{Breiman2001}
  \item Support Vector Regression (SVR) \citep{CortesVapnik1995,ChangLin2011}
  \item Extreme Gradient Boosting (XGBoost) \citep{ChenGuestrin2016}
  \item TabNet \citep{Arik2021}
\end{itemize}

All models were trained using identical feature sets and evaluation protocols.
Based on predictive accuracy and stability across random seeds, Random Forest and Multilayer Perceptron were selected for all subsequent analyses. These two models represent complementary inductive biases: RF captures non-linear interactions through ensemble tree structures, while MLP provides a flexible function approximation via neural networks.

To ensure consistency and reproducibility, only RF and MLP are reported in all experiments and tables.

\subsection{Model Training and Predictive Validity}

Models were trained on elite football data using standard regression objectives to predict match outcomes. Predictive performance was evaluated on a held-out elite test set using mean absolute error (MAE) and root mean squared error (RMSE). Each model was trained multiple times with different random seeds to assess performance stability.

Only models demonstrating stable predictive performance on elite data were considered for interpretability analysis. Predictive metrics on university football are not reported, as the study does not aim to evaluate generalization accuracy across domains.

\subsection{Explanation Methods}

To evaluate interpretability robustness, two conceptually distinct explanation methods were employed. Let $f(\cdot)$ denote the trained prediction model used throughout this section. For implementation, SHAP explanations use TreeExplainer for RF and DeepExplainer for MLP, and all experiments are repeated across five random seeds (0, 1, 2, 3, and 4).

\subsubsection{SHapley Additive exPlanations (SHAP)}

SHapley Additive exPlanations (SHAP) provide game-theoretic feature attributions by estimating each feature's marginal contribution to model predictions \citep{LundbergLee2017}. For an input sample $x$ with $p$ features, the SHAP value for feature $j$ is defined as:

\begin{equation}
\phi_j(x)=\sum_{S\subseteq F\setminus\{j\}}\frac{|S|!(p-|S|-1)!}{p!}
\left[f(S\cup\{j\})-f(S)\right],
\end{equation}

where $F=\{1,\ldots,p\}$ is the full feature set, $S$ denotes a feature subset that excludes $j$. Thus, $f(S\cup\{j\})-f(S)$ is the marginal contribution of feature $j$.

In this study, global SHAP importance for feature $j$ on domain $\mathcal{D}$ is computed as the mean absolute attribution:

\begin{equation}
I^{\mathrm{SHAP}}_j(\mathcal{D})=\frac{1}{N_{\mathcal{D}}}\sum_{n=1}^{N_{\mathcal{D}}}\left|\phi_j(x_n)\right|,
\end{equation}

where $N_{\mathcal{D}}$ is the number of samples in domain $\mathcal{D}$. Larger values indicate that the model relies more strongly on that feature for prediction. Instability of SHAP rankings across seeds or domains suggests context-dependent feature relevance rather than causal effect.

\subsubsection{Counterfactual Impact Score (CIS)}

The Counterfactual Impact Score (CIS) estimates feature importance by measuring prediction changes under controlled single-feature perturbations. Let $x_n \in \mathbb{R}^{p}$ denote the $n$-th sample in an evaluation domain $\mathcal{D}$ with $p$ features. For each feature $j$, a directional target value is defined from elite-training statistics:

\begin{equation}
t_j=\mu_j^{\text{elite}}+k_j\,\sigma_j^{\text{elite}},
\end{equation}

where $\mu_j^{\text{elite}}$ and $\sigma_j^{\text{elite}}$ are the elite-training mean and standard deviation, and $k_j \in \{+1,-1\}$ controls perturbation direction. We set $k_j=-1$ for failure-type features (feature name contains ``fail'' or equals ``offside''/``fouls'') and $k_j=+1$ otherwise. For decrease perturbations ($k_j=-1$), $t_j$ is lower-bounded at 0.

The counterfactual sample $x_n^{(j)}$ is constructed by changing only feature $j$ toward $t_j$:

\begin{equation}
x_{n,\ell}^{(j)}=
\begin{cases}
\max(x_{n,j},t_j), & \ell=j \text{ and } k_j=+1, \\
\min(x_{n,j},t_j), & \ell=j \text{ and } k_j=-1, \\
x_{n,\ell}, & \ell\neq j.
\end{cases}
\end{equation}

The unnormalized CIS for feature $j$ on domain $\mathcal{D}$ is then defined as:

\begin{equation}
\mathrm{CIS}_j(\mathcal{D})=\frac{1}{N_{\mathcal{D}}}\sum_{n=1}^{N_{\mathcal{D}}}\left|f\!\left(x_n^{(j)}\right)-f(x_n)\right|,
\end{equation}

where $N_{\mathcal{D}}$ is the number of samples in domain $\mathcal{D}$. To facilitate cross-feature and cross-domain comparison, CIS values are normalized as:

\begin{equation}
\widetilde{\mathrm{CIS}}_j(\mathcal{D})=
\frac{\mathrm{CIS}_j(\mathcal{D})}{\sum_{k=1}^{p}\mathrm{CIS}_k(\mathcal{D})},
\end{equation}

such that $\sum_{j=1}^{p}\widetilde{\mathrm{CIS}}_j(\mathcal{D})=1$.

Higher CIS values indicate that a one-standard-deviation directional shift in a feature induces a larger change in predicted performance, suggesting higher sensitivity. Conversely, low CIS values imply minimal impact on predictions, indicating relative insensitivity. Unlike SHAP, CIS captures sensitivity to directional feature shifts rather than frequency of feature usage in attribution.

\subsubsection{Reproducibility}


For reproducibility, SHAP uses TreeExplainer for RF and DeepExplainer for MLP, with a fixed MLP background subset from elite training that is reused across evaluated domains within each seed. Elite training data are used exclusively to construct SHAP backgrounds and CIS reference statistics (no university samples are used). Five seeds (\{0,1,2,3,4\}) are applied consistently for model training and SHAP background sampling, while CIS is deterministic given a trained model.

\subsection{Explanation Stability, Robustness, and Statistical Analysis}

Interpretability robustness was evaluated along three dimensions:

\begin{enumerate}
  \item \textbf{Seed-wise stability:} Feature importance rankings were compared across multiple random seeds using rank correlation metrics.
  \item \textbf{Domain-wise transferability:} Explanation structures were compared between elite leagues and university football to assess structural alignment.
  \item \textbf{Method agreement:} Feature importance rankings obtained via SHAP and CIS were compared to evaluate method invariance.
\end{enumerate}

Explanation variability is interpreted as reflecting structural ambiguity in the target domain.
To quantify robustness, rank-based Spearman correlation coefficients were used to measure agreement between feature importance rankings across seeds, domains, and explanation methods. Statistical significance was evaluated at $\alpha = 0.05$ unless otherwise specified. Results are reported as mean values with corresponding variability measures to reflect robustness.

\section{Results}
\subsection{Experiment 1: Predictive Validity on Elite Football Data}

\begin{table}[h]
\tbl{Predictive performance of machine learning models on elite football data (elite test set).
Values are reported as mean $\pm$ standard deviation across five random seeds.}
{\begin{tabular*}{0.8\linewidth}{@{\extracolsep{\fill}}lcc}
\toprule
\textbf{Method} & \textbf{MAE (Test)} & \textbf{RMSE (Test)} \\
\midrule
DT     & $1.771 \pm 0.021$ & $2.318 \pm 0.036$ \\
\textbf{MLP}    & $1.180 \pm 0.009$ & $1.550 \pm 0.014$ \\
\textbf{RF}     & $1.202 \pm 0.005$ & $1.560 \pm 0.004$ \\
SVR    & $1.215 \pm 0.000$ & $1.599 \pm 0.000$ \\
XGB    & $1.253 \pm 0.000$ & $1.625 \pm 0.000$ \\
TabNet & $1.209 \pm 0.007$ & $1.585 \pm 0.013$ \\
\bottomrule
\end{tabular*}}
\label{tab:predictive_validity}
\end{table}

The first experiment evaluates whether the selected machine learning models achieve stable predictive performance on elite football data, establishing a foundation for subsequent interpretability analysis.

Models were trained on elite football data and evaluated on a held-out elite test set using mean absolute error (MAE) and root mean squared error (RMSE). To assess robustness with respect to training randomness, each model was trained multiple times using different random seeds. Results are summarized in Table~\ref{tab:predictive_validity} as mean values with corresponding standard deviations.

As shown in Table~\ref{tab:predictive_validity}, RF and MLP achieved the lowest error values with minimal variability across random seeds, indicating stable predictive behavior. DT, SVR, XGB, and TabNet exhibited higher error levels or greater variability, suggesting reduced robustness under the given feature representation.

Based on these results, RF and MLP were selected for subsequent explanation analysis, representing tree-based and neural network paradigms, respectively.

\subsection{Experiment 2: Feature Importance Comparison Across Domains}

Experiment 2 evaluates whether performance determinants learned from elite football are structurally transferable to university football under a consistent feature space. Global feature importance was computed separately for elite test data and \targetuniversitydata{} using SHAP and CIS under RF and MLP. Feature importance was normalized within each domain to enable comparison of relative rankings.

\subsubsection{Comparison for RF}

\begin{figure}[h]
\centering
\subfloat[RF | SHAP: Elite vs. \Targetdomainshort{}]{%
\includegraphics[width=0.48\linewidth]{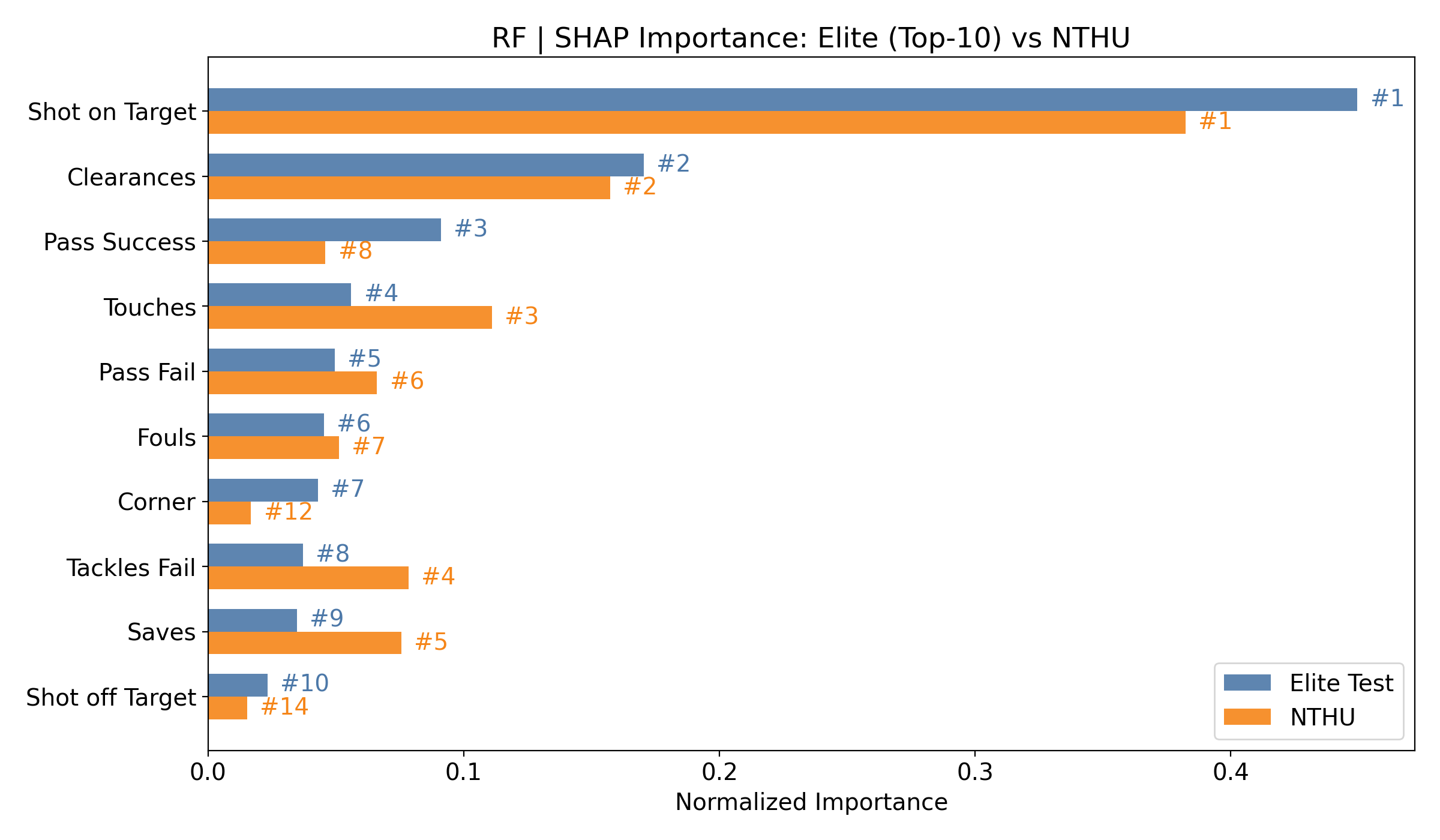}}
\hfill
\subfloat[RF | CIS: Elite vs. \Targetdomainshort{}]{%
\includegraphics[width=0.48\linewidth]{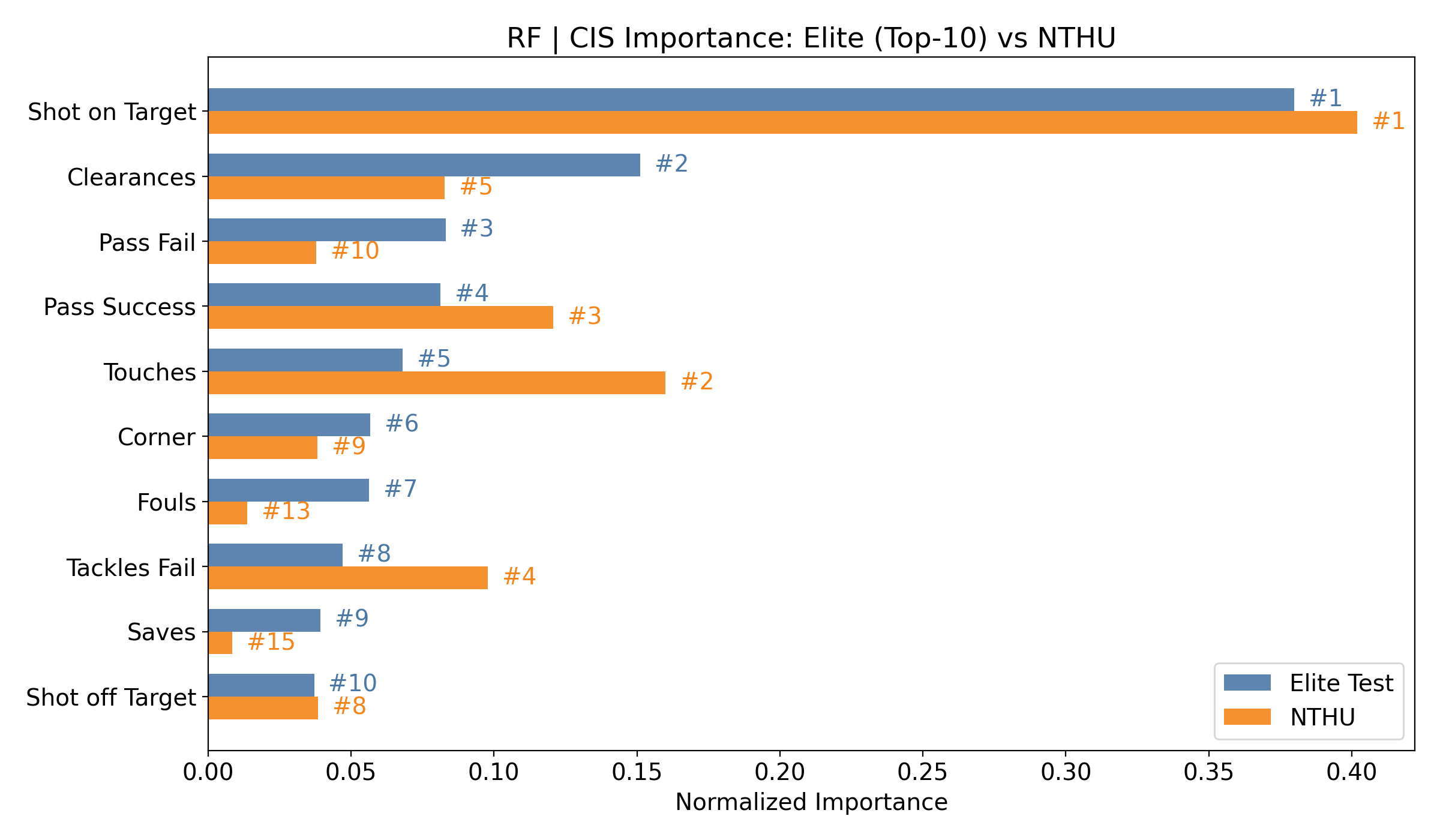}}
\caption{Global feature importance comparison between elite football and \targetdomainlong{} using RF. SHAP and CIS provide complementary importance estimates; feature rankings differ across domains despite identical feature definitions.}
\label{fig:exp2_rf_importance}
\end{figure}

Figure~\ref{fig:exp2_rf_importance} presents the global feature importance rankings derived from the RF model. In the elite domain, feature importance is concentrated among a small subset of indicators, forming a coherent hierarchical structure. In contrast, the \targetdomainshort{} domain exhibits a different ordering of influential features, with importance distributed more evenly and several indicators shifting relative positions. For example, under CIS, \textit{Touches} rises from rank 5 (elite) to rank 2 (\targetdomainshort{}). This shift suggests that the RF explanation assigns greater importance to possession-related actions in the university domain.

Both SHAP and CIS consistently reveal domain-dependent differences in feature hierarchy under the RF model.

\subsubsection{Comparison for MLP}

\begin{figure}[h]
\centering
\subfloat[MLP | SHAP: Elite vs. \Targetdomainshort{}]{%
\includegraphics[width=0.48\linewidth]{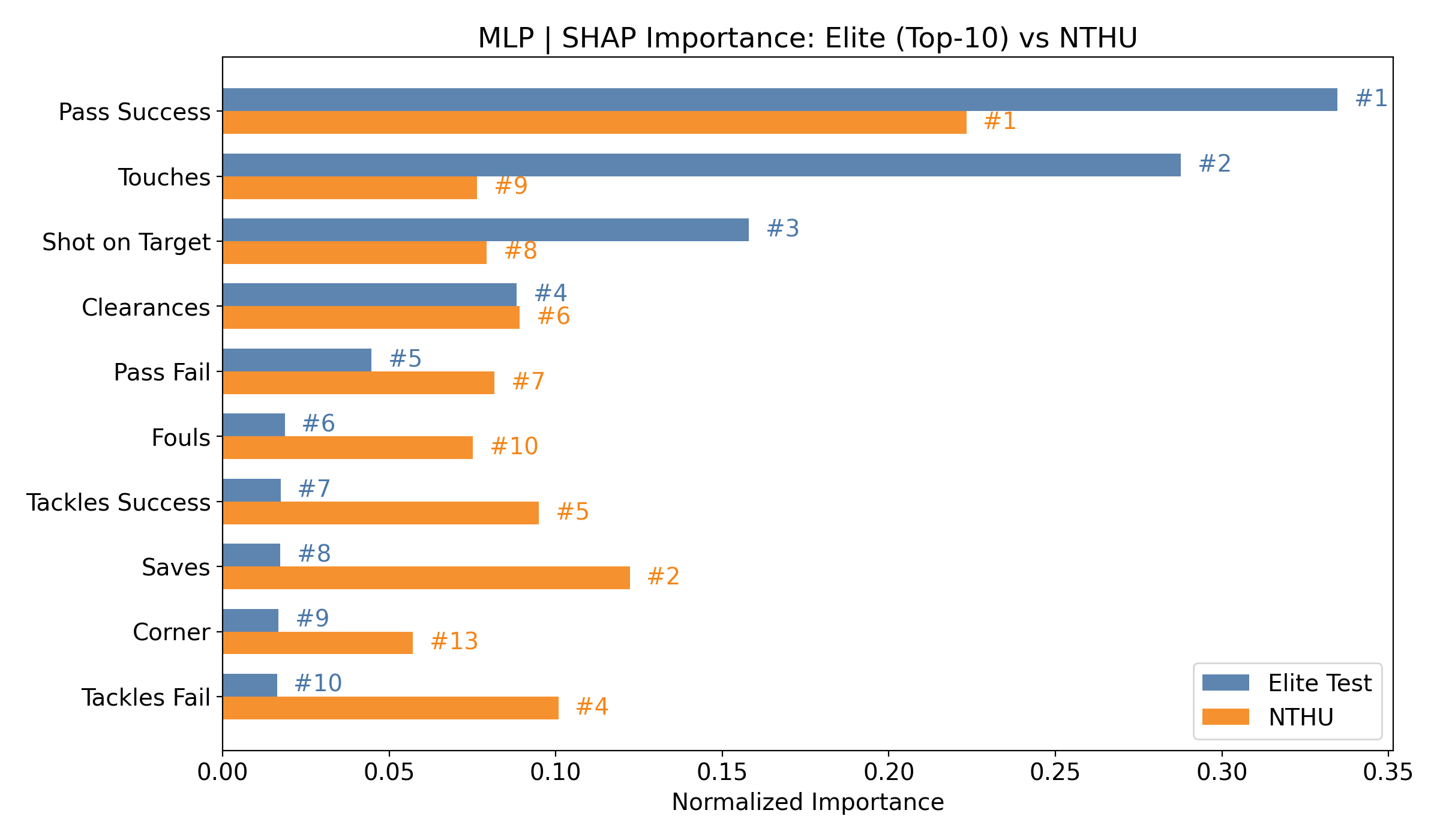}}
\hfill
\subfloat[MLP | CIS: Elite vs. \Targetdomainshort{}]{%
\includegraphics[width=0.48\linewidth]{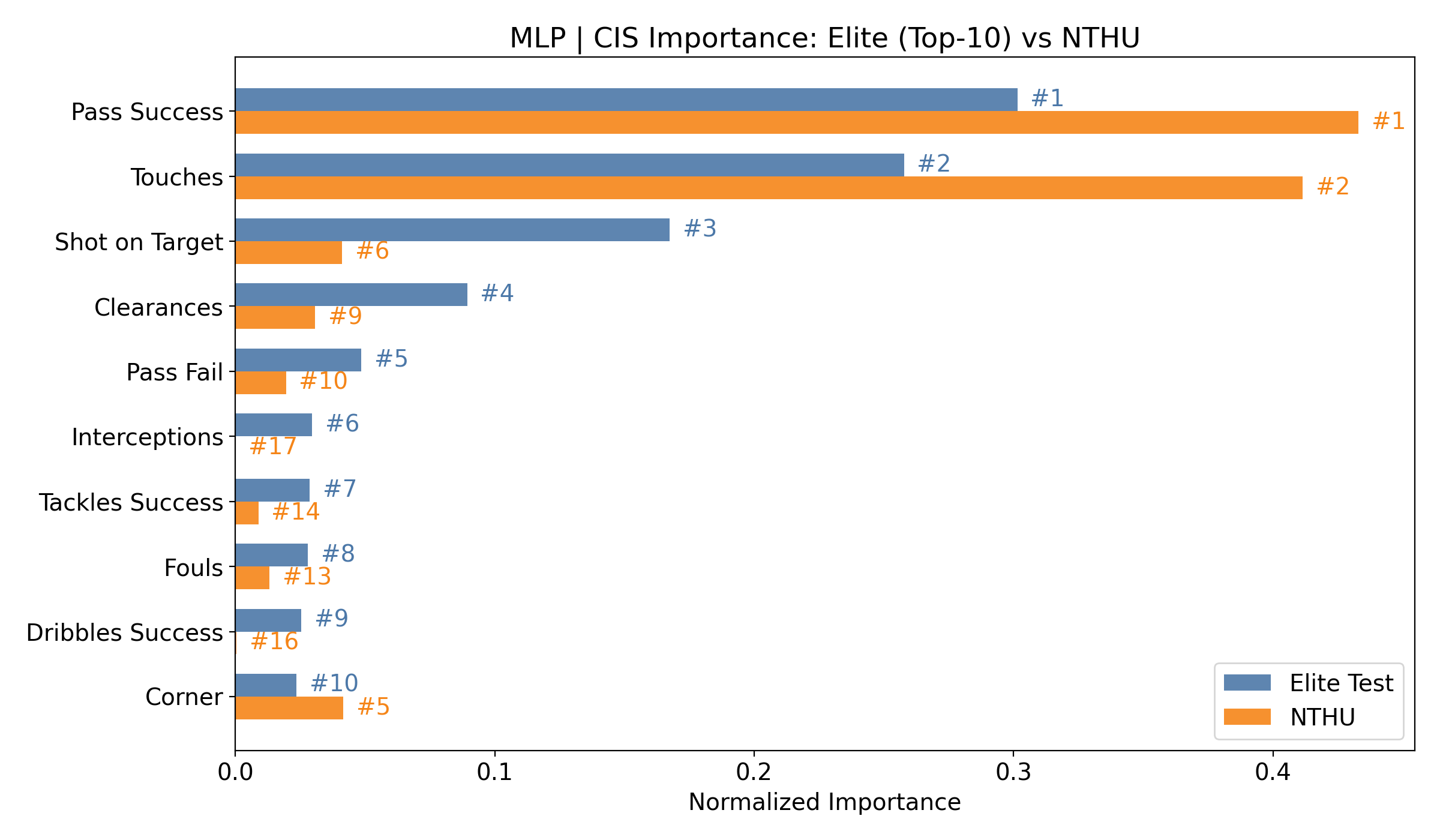}}
\caption{Global feature importance comparison between elite football and \targetdomainlong{} using MLP. SHAP and CIS reveal domain-dependent reordering of important indicators under identical feature space.}
\label{fig:exp2_mlp_importance}
\end{figure}

Figure~\ref{fig:exp2_mlp_importance} shows corresponding results for the MLP model. Similar patterns are observed: the elite domain demonstrates a concentrated importance structure, while the \targetdomainshort{} domain exhibits reordering among top-ranked features and a flatter importance distribution. For example, under SHAP, \textit{Saves} rises from rank 8 (elite) to rank 2 (\targetdomainshort{}). This suggests that the MLP explanation places greater importance on defensive goal-prevention events in the university domain.

This divergence persists across both SHAP and CIS, indicating that the effect is not model-specific.

\subsubsection{Cross-model and cross-method consistency.}

Across both models and explanation methods, the elite and \targetdomainshort{} domains consistently exhibit different feature importance structures. This suggests that the observed divergence reflects systematic differences in performance determinant organization rather than artifacts of a specific model or explanation technique.

Taken together, these results indicate that performance determinants learned from elite football are not structurally preserved when applied to university football.

\subsection{Experiment 3: Seed-wise Explanation Reproducibility}

\begin{figure}[h]
\centering
\subfloat[RF]{%
\includegraphics[width=0.48\linewidth]{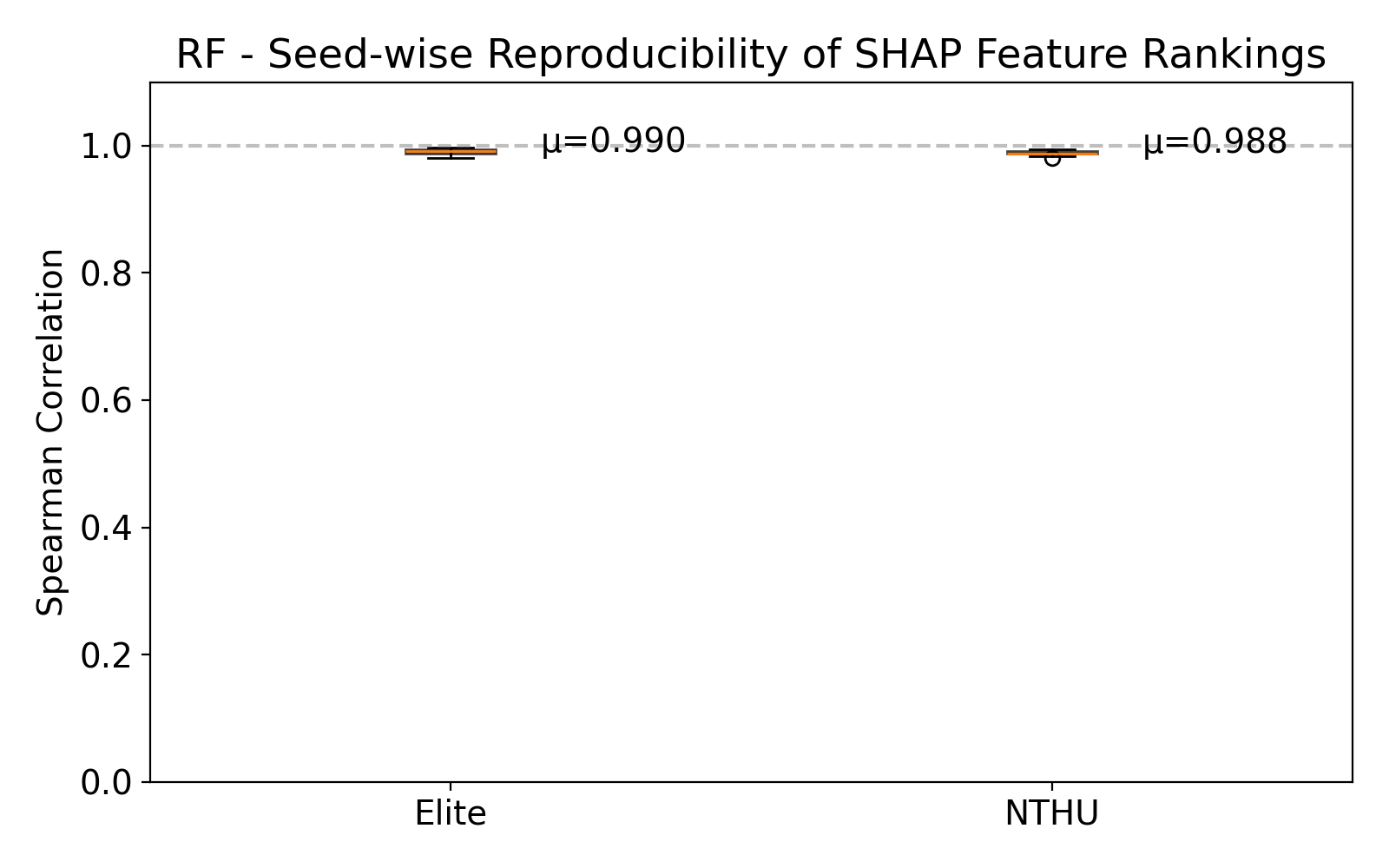}}
\hfill
\subfloat[MLP]{%
\includegraphics[width=0.48\linewidth]{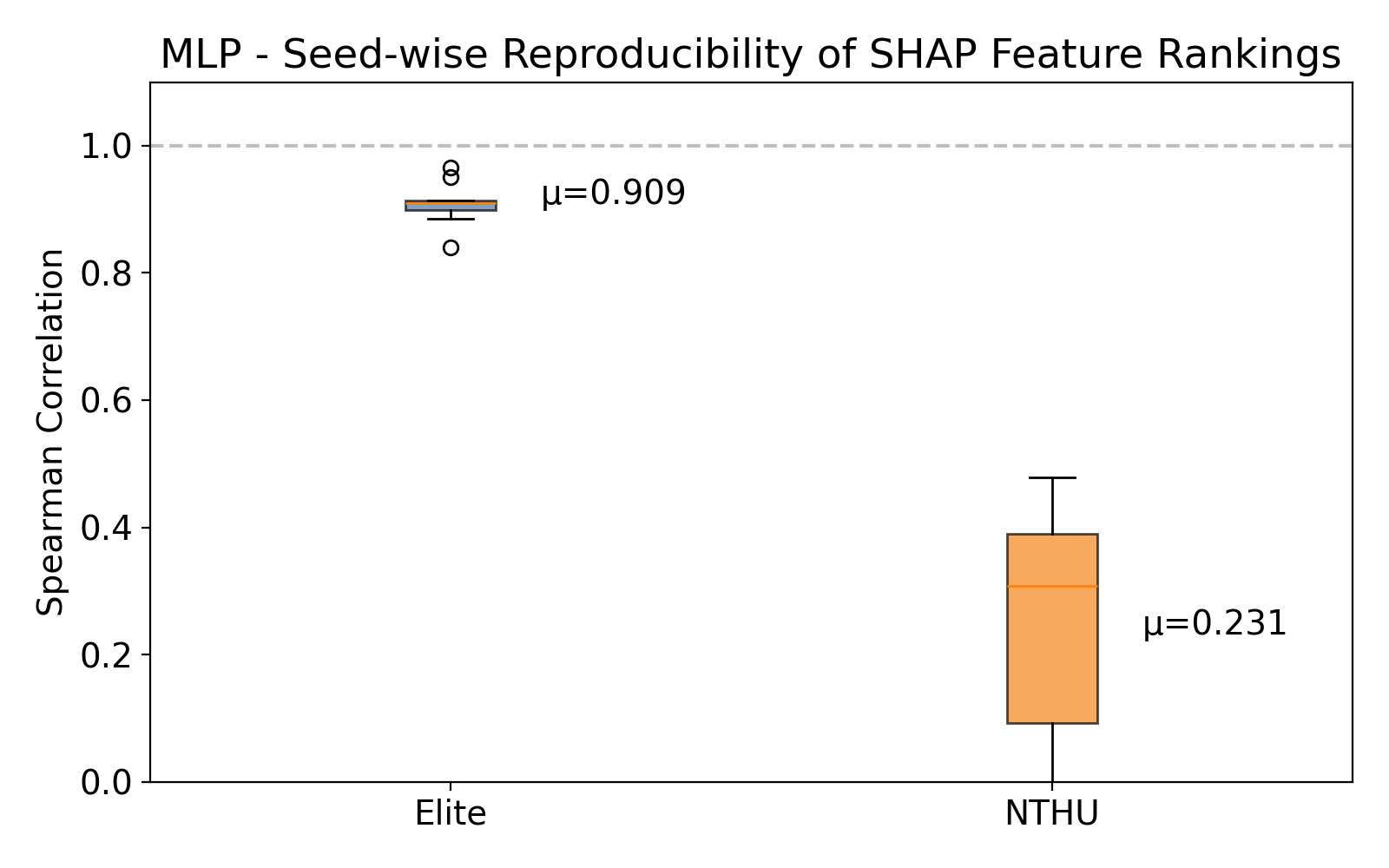}}
\caption{Seed-wise reproducibility of SHAP feature importance rankings measured by pairwise Spearman rank correlation. Distributions are shown separately for elite football and \targetdomainlong{} across multiple random seeds.}
\label{fig:exp3_stability}
\end{figure}

Experiment~3 evaluates the reproducibility of feature importance rankings under training randomness. Following Experiment~2, this analysis examines whether explanation structures remain consistent across different random seeds.

For each model, SHAP-based feature importance rankings were computed across multiple training runs, and pairwise Spearman rank correlations were used to quantify reproducibility. The resulting distributions are summarized in Figure~\ref{fig:exp3_stability}.

\subsubsection{Elite football domain}
In the elite domain, feature importance rankings exhibit high reproducibility across random seeds. Quantitatively, the mean pairwise Spearman correlation is $\mu=0.990$ for RF and $\mu=0.909$ for MLP, indicating strong agreement of feature rankings across runs.

\subsubsection{University football domain}
In the university domain, model-specific differences become clearer. RF remains highly stable ($\mu=0.988$), whereas MLP drops to $\mu=0.231$ and shows much wider dispersion (approximately 0.000--0.485), indicating unstable seed-wise rankings for the neural model.

\subsubsection{Cross-domain comparison}
Cross-domain reproducibility is therefore model-dependent: RF is consistently stable in both domains, while MLP shows a pronounced decline from elite to \targetdomainshort{} ($0.91 \rightarrow 0.23$). This pattern indicates that domain shift can substantially amplify explanation instability for some model classes.

Taken together, these results demonstrate that explanation stability is both domain- and model-dependent.

\subsection{Experiment 4: Structural Agreement Across Domains}

\begin{figure}[h]
\centering
\subfloat[Random Forest (RF)]{%
\includegraphics[width=0.48\linewidth]{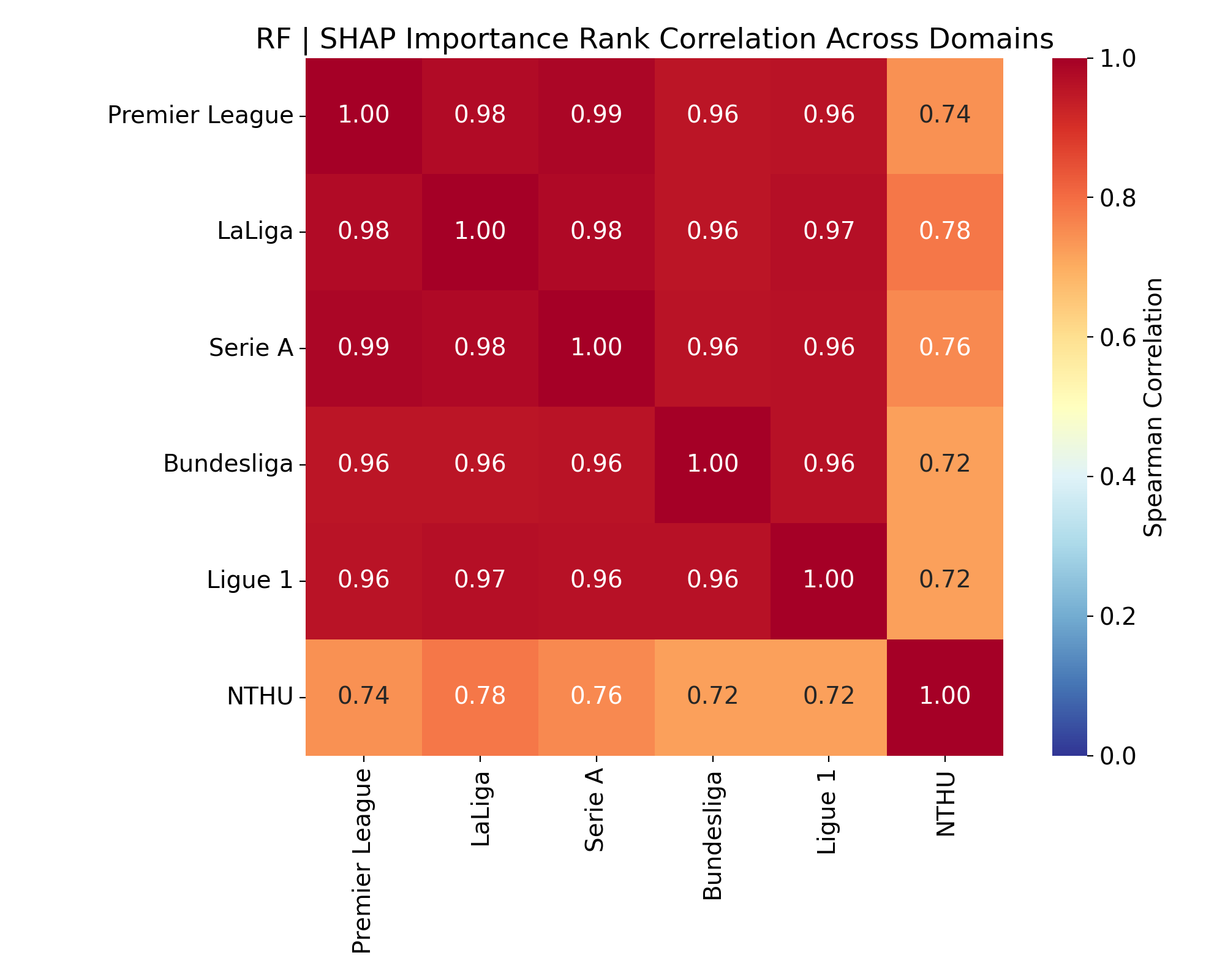}}
\hfill
\subfloat[Multilayer Perceptron (MLP)]{%
\includegraphics[width=0.48\linewidth]{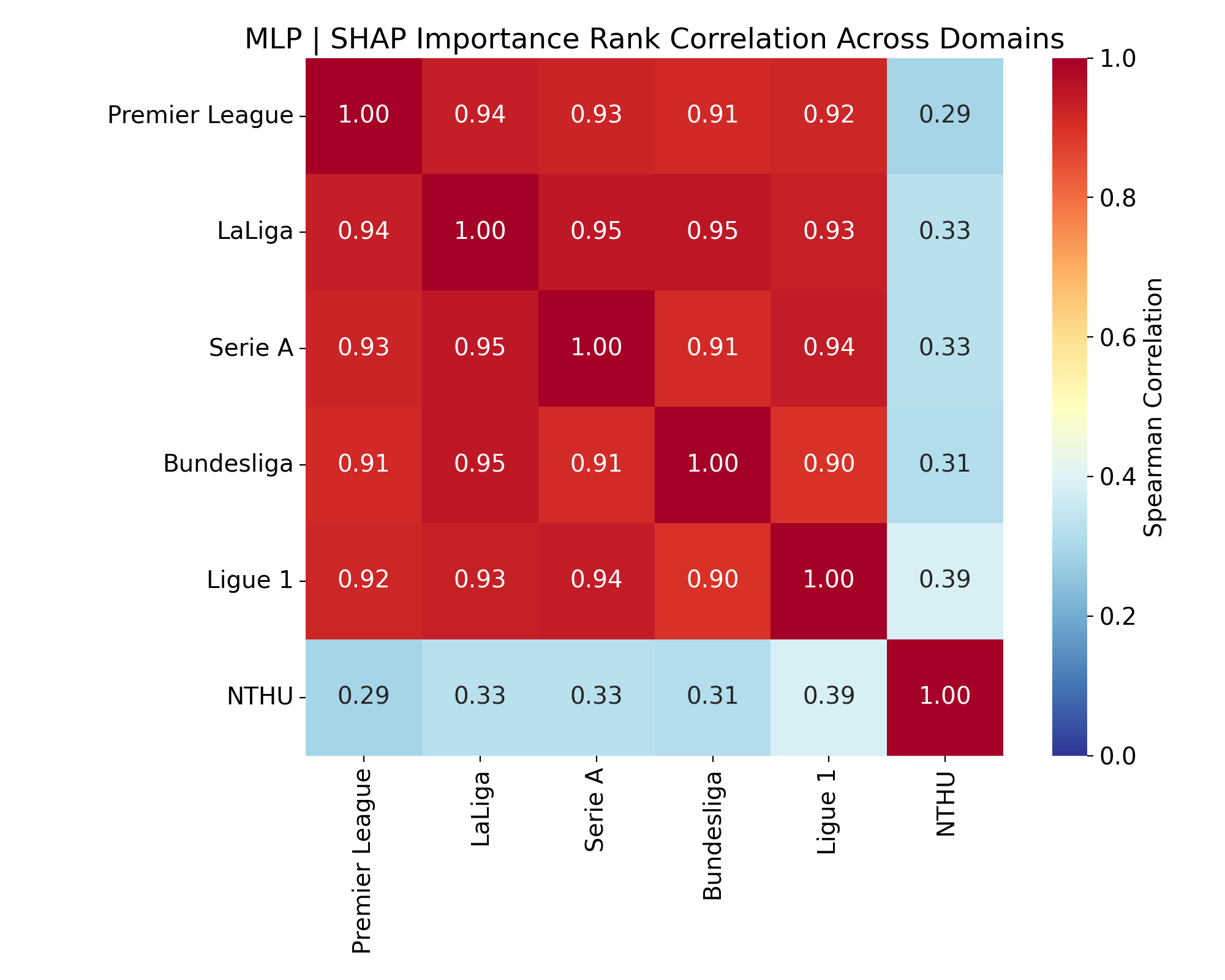}}
\caption{SHAP-based structural agreement of feature importance rankings across domains. Each cell represents the Spearman rank correlation between a pair of domains. Strong agreement is observed among elite leagues, while correlations between elite leagues and \targetdomainlong{} are substantially lower.}
\label{fig:exp4_shap_heatmap}
\end{figure}

Experiment~4 evaluates whether football performance determinants share a common structural organization across competition levels. Each elite league and the \targetuniversitydataset{} were treated as separate domains, and global feature importance rankings were computed for each domain under the same trained model. Pairwise Spearman rank correlations were then calculated to quantify structural agreement across domains.

Figure~\ref{fig:exp4_shap_heatmap} presents the SHAP-based correlation heatmap. Strong agreement is observed among elite leagues, with consistently high correlation values indicating a shared importance structure. 

In contrast, correlations between elite leagues and \targetdomainlong{} are substantially lower, indicating a lack of structural alignment. This pattern is consistent across multiple elite leagues rather than driven by a single comparison.

The observed pattern of strong elite--elite agreement and weak elite--university agreement is consistent across both RF and MLP models. Although absolute correlation values vary slightly, the overall structure of the heatmaps remains unchanged.

Taken together, these results indicate that performance determinants exhibit a shared structure across elite competitions but diverge at the university level.

\subsection{Experiment 5: Explanation-Method Agreement}

\begin{table}[h]
\tbl{Agreement between SHAP and CIS feature importance rankings
measured by Spearman rank correlation.
Elite-average values are computed by averaging across the top five European leagues.
Correlation values are reported as mean $\pm$ standard deviation across five random seeds,
with median $p$-values shown in parentheses.}
{\begin{tabular*}{0.8\linewidth}{@{\extracolsep{\fill}}lcc}
\toprule
\textbf{Domain} & \textbf{RF} & \textbf{MLP} \\
\midrule
Elite
& $0.932 \pm 0.019$ $(p < 10^{-6})$
& $0.826 \pm 0.050$ $(p < 10^{-4})$ \\

\Targetdomainshort{}
& $0.420 \pm 0.018$ $(p = 0.10)$
& $0.323 \pm 0.178$ $(p = 0.24)$ \\
\bottomrule
\end{tabular*}}
\label{tab:method_agreement}
\end{table}

Experiment~5 evaluates whether feature importance rankings are consistent across explanation methods within the same domain. Specifically, agreement between SHAP and CIS is assessed under identical model outputs.

For each domain, feature importance rankings were computed using both SHAP and CIS. Agreement was quantified using Spearman rank correlation across multiple random seeds. Results are summarized in Table~\ref{tab:method_agreement} as mean values with corresponding standard deviations, along with median $p$-values.

As shown in Table~\ref{tab:method_agreement}, elite football domains exhibit consistently high agreement between SHAP and CIS rankings for both RF and MLP models. Correlation values are strong and statistically significant, indicating that feature importance rankings are largely consistent across explanation methods in elite competitions.

In contrast, agreement between SHAP and CIS decreases substantially in the \targetdomainlong{} domain. Correlation values are lower, exhibit greater variability, and are not consistently statistically significant. This indicates that feature importance rankings become dependent on the choice of explanation method in the university domain.

Taken together, these results demonstrate that explanation-method agreement is domain-dependent, with consistent interpretations in elite football but increased method sensitivity in university football.

\section{Discussion}
This paper examined whether football performance determinants learned from elite competitions are structurally transferable to university-level football, and under what conditions model explanations remain trustworthy under domain shift. Across five experiments, the results show that explanation stability, agreement, and transferability depend strongly on competition level, with important implications for applied football analytics.

\subsection{Structural transferability of performance determinants}

Results from Experiments~2 and~4 provide convergent evidence that football performance determinants exhibit a shared structural organization across elite competitions but diverge  when applied to university football. 

Feature-importance rankings derived from elite leagues show strong cross-domain alignment, indicating a coherent performance structure across elite competitions. This finding is consistent with prior work suggesting that elite football displays relatively stable tactical and performance patterns across leagues and seasons.

In contrast, the same feature space does not preserve its structural hierarchy when evaluated on \targetdomainlong{}. Both SHAP- and CIS-based analyses reveal substantial reordering of key indicators and reduced cross-domain agreement. This divergence persists across models with different inductive biases, providing evidence consistent with domain-level structural differences rather than purely model-specific effects.

However, alternative explanations, including limited target-domain sample size, site specificity, and annotation variability, cannot be fully excluded. These findings suggest that performance determinants in university football may follow a different structure, rather than just being weaker versions of elite-level patterns.

\subsection{Interpretability stability as a domain-dependent property}

Experiment~3 demonstrates that explanation reproducibility under random initialization varies markedly across domains. While explanations derived from elite football remain highly stable across random seeds, those obtained from university football---particularly under more expressive models---exhibit substantially greater variability. This highlights that explanation stability should not be assumed \textit{a priori}, especially when models are applied outside their training domain.

Rather than indicating methodological failure, the observed instability reflects the absence of a consolidated importance hierarchy, such that small variations in training conditions lead to divergent explanatory outcomes. In this sense, instability itself is informative, signaling that learned representations do not converge toward a single dominant structure. These results suggest that interpretability should be viewed as a domain-dependent property reflecting the degree of structural consistency in the underlying data.

\subsection{Explanation-method agreement and interpretability trustworthiness}

Experiment~5 examines whether feature importance rankings remain consistent across explanation methods. In elite football, SHAP and CIS produce highly similar rankings with strong statistical support, indicating consistent interpretations across different explanation approaches.

In contrast, agreement between SHAP and CIS decreases substantially in the university domain, with lower correlations and inconsistent statistical significance across random seeds. This method dependence indicates that feature importance rankings are sensitive to the choice of explanation method, suggesting the absence of a stable underlying importance structure. From an applied perspective, this finding cautions against relying on a single explanation method to derive actionable insights in such settings.

\subsection{Implications for applied football analytics}

The findings carry several implications for practitioners. First, explanations derived from elite football models should not be directly transferred to university or amateur contexts, even when feature definitions are identical. Structural divergence implies that indicators important at the elite level may not play equivalent roles at lower competition levels.

Second, explanation instability and method disagreement should be treated as diagnostic signals rather than noise. Unstable explanations may reflect tactical ambiguity, developmental heterogeneity, or inconsistent execution, indicating that performance structures have not yet stabilized.

Within this framework, SHAP and CIS provide complementary insights. SHAP reflects features most relied upon by the model, while CIS captures sensitivity to deviations from elite-level norms. Features exhibiting both stable SHAP importance and high CIS may be interpreted as structurally salient indicators, whereas features with inconsistent rankings suggest context-dependent effects requiring cautious interpretation.

Overall, these findings highlight the importance of evaluating interpretability robustness—across seeds, models, methods, and domains—before applying explanation-based insights in practice.

\subsection{Limitations and future directions}

Several limitations should be acknowledged. First, the university football data originate from a single institution with a limited sample size, which may constrain generalizability across collegiate contexts. In addition, university football is not a homogeneous competitive environment, as teams from different divisions may differ substantially in playing strength and competitive demands. Future research could therefore extend this framework to multiple university programs, divisions, or larger datasets to examine whether similar structural patterns emerge under a broader range of collegiate conditions.

In addition, this study focuses on global feature importance. Future work could investigate local or context-specific explanations to assess whether interpretability stability varies across match phases or tactical scenarios. Incorporating spatio-temporal representations may further improve the characterization of performance structures across competition levels.




\section{Conclusion}

This paper investigated whether football performance determinants learned from elite competitions are structurally transferable to university-level football and under what conditions model explanations can be considered trustworthy. By systematically analyzing predictive validity, feature importance structure, explanation stability, cross-domain agreement, and explanation-method consistency, this work shifts the focus from model performance to interpretability as an object of empirical evaluation.

The results indicate that elite football exhibits a stable and shared hierarchy of performance determinants across leagues, models, and explanation methods. In contrast, university football shows substantially weaker structural alignment, reduced explanation stability, and increased sensitivity to explanation choice. These findings are consistent with domain-level structural differences rather than a simple noisier version of elite indicators.

Importantly, explanation instability and method disagreement are not interpreted as methodological failures. Instead, they serve as diagnostic signals of structural ambiguity under domain shift. These findings highlight that interpretability should be treated as a conditional and data-dependent property rather than an existing characteristic of machine learning models.

Overall, this paper underscores the necessity of evaluating explanation robustness before deploying interpretability-based insights in applied football analytics. By framing interpretability itself as a measurable and informative outcome, the study provides a principled foundation for responsible use of machine learning explanations across different levels of football competition.

\section*{Acknowledgments}
The authors thank the National Tsing Hua University football players and Coach Kok-Hua Tan for their support and assistance throughout this study.
Generative AI tools (ChatGPT with version GPT-5.4) were used solely for English grammar correction and manuscript polishing. No generative AI tools were used to generate the research content, conduct the analyses, or produce the results, and the authors take full responsibility for the manuscript.

\section*{Declaration of interest statement}
No potential conflict of interest was reported by the author(s).

\section*{Funding details}
This research received no external funding.

\section*{Data availability statement}
The elite dataset is publicly available from WhoScored.com and is annotated according to Opta event definitions. The \targetdataset{} is not publicly available due to privacy considerations and institutional restrictions on data sharing.

\bibliographystyle{apalike}
\bibliography{references}

@inproceedings{Arik2021,
  author    = {Arik, S. O. and Pfister, T.},
  title     = {TabNet: Attentive interpretable tabular learning},
  booktitle = {Proceedings of the AAAI Conference on Artificial Intelligence},
  year      = {2021},
  volume    = {35},
  pages     = {6679--6687}
}

@article{AttaMills2024,
  author  = {Atta Mills, E. F. E. and Deng, Z. and Zhong, Z. and Li, J.},
  title   = {Data-driven prediction of soccer outcomes using enhanced machine and deep learning techniques},
  journal = {Journal of Big Data},
  year    = {2024},
  volume  = {11},
  pages   = {170}
}

@article{Berrar2024,
  author  = {Berrar, D. and Lopes, P. and Dubitzky, W.},
  title   = {A data- and knowledge-driven framework for developing machine learning models to predict soccer match outcomes},
  journal = {Machine Learning},
  year    = {2024},
  volume  = {113},
  pages   = {8165--8204}
}

@article{Biermann2025,
  author  = {Biermann, H. and Memmert, D. and Petersen, N. and Raabe, D.},
  title   = {Contextualization of soccer analysis with tactical periodization and machine learning},
  journal = {Data Mining and Knowledge Discovery},
  year    = {2025},
  volume  = {39},
  pages   = {23}
}

@book{Breiman1984,
  author    = {Breiman, L. and Friedman, J. and Stone, C. J. and Olshen, R. A.},
  title     = {Classification and Regression Trees},
  publisher = {CRC Press},
  year      = {1984}
}

@article{Breiman2001,
  author  = {Breiman, L.},
  title   = {Random forests},
  journal = {Machine Learning},
  year    = {2001},
  volume  = {45},
  pages   = {5--32}
}

@book{Carling2007,
  author    = {Carling, C. and Williams, A. M. and Reilly, T.},
  title     = {Handbook of soccer match analysis: A systematic approach to improving performance},
  publisher = {Routledge},
  year      = {2007}
}

@article{Cavus2023,
  author  = {Cavus, Mustafa and Stando, Adrian and Biecek, Przemyslaw},
  title   = {Glocal Explanations of Expected Goal Models in Soccer},
  journal = {CoRR},
  year    = {2023},
  volume  = {abs/2308.15559},
  doi     = {10.48550/arXiv.2308.15559}
}

@article{ChangLin2011,
  author  = {Chang, C.-C. and Lin, C.-J.},
  title   = {{LIBSVM}: A library for support vector machines},
  journal = {ACM Transactions on Intelligent Systems and Technology},
  year    = {2011},
  volume  = {2},
  number  = {3},
  pages   = {1--27}
}

@inproceedings{ChenGuestrin2016,
  author    = {Chen, T. and Guestrin, C.},
  title     = {{XGBoost}: A scalable tree boosting system},
  booktitle = {Proceedings of the 22nd ACM SIGKDD International Conference on Knowledge Discovery and Data Mining},
  year      = {2016},
  pages     = {785--794}
}

@article{CortesVapnik1995,
  author  = {Cortes, C. and Vapnik, V.},
  title   = {Support-vector networks},
  journal = {Machine Learning},
  year    = {1995},
  volume  = {20},
  pages   = {273--297}
}

@inproceedings{Decroos2019,
  author    = {Decroos, T. and Bransen, L. and Van Haaren, J. and Davis, J.},
  title     = {Actions Speak Louder Than Goals: Valuing Player Actions in Soccer},
  booktitle = {Proceedings of the 25th ACM SIGKDD International Conference on Knowledge Discovery \& Data Mining},
  year      = {2019},
  pages     = {1851--1861},
  doi       = {10.1145/3292500.3330758}
}

@article{Franca2022,
  author  = {Fran{\c{c}}a, C. and Ihle, A. and Marques, A. and Sarmento, H. and Martins, F. and Henriques, R. and Gouveia, {\'E}. R.},
  title   = {Physical development differences between professional soccer players from different competitive levels},
  journal = {Applied Sciences},
  year    = {2022},
  volume  = {12},
  number  = {14},
  pages   = {7343}
}

@article{GarciaAliaga2021,
  author  = {Garc{\'i}a-Aliaga, A. and Marquina, M. and Coteron, J. and Rodr{\'i}guez-Gonz{\'a}lez, A. and Luengo-Sanchez, S.},
  title   = {In-game behaviour analysis of football players using machine learning techniques based on player statistics},
  journal = {International Journal of Sports Science \& Coaching},
  year    = {2021},
  volume  = {16},
  number  = {1},
  pages   = {148--157}
}

@article{GomezRuano2020,
  author  = {G{\'o}mez-Ruano, M. A. and Ib{\'a}{\~n}ez, S. J. and Leicht, A. S.},
  title   = {Performance analysis in sport},
  journal = {Frontiers in Psychology},
  year    = {2020},
  volume  = {11},
  pages   = {611634}
}

@incollection{Javed2023,
  author    = {Javed, D. and Jhanjhi, N. Z. and Khan, N. A.},
  title     = {Football analytics for goal prediction to assess player performance},
  booktitle = {Innovation and Technology in Sports},
  publisher = {Springer},
  address   = {Singapore},
  year      = {2023},
  pages     = {245--257}
}

@article{Kusmakar2020,
  author  = {Kusmakar, S. and Shelyag, S. and Zhu, Y. and Dwyer, D. and Gastin, P. and Angelova, M.},
  title   = {Machine learning enabled team performance analysis in the dynamical environment of soccer},
  journal = {IEEE Access},
  year    = {2020},
  volume  = {8},
  pages   = {90266--90279}
}

@article{Li2020,
  author  = {Li, Y. and Ma, R. and Gon{\c{c}}alves, B. and Gong, B. and Cui, Y. and Shen, Y.},
  title   = {Data-driven team ranking and match performance analysis in the Chinese Football Super League},
  journal = {Chaos, Solitons \& Fractals},
  year    = {2020},
  volume  = {141},
  pages   = {110330}
}

@inproceedings{LundbergLee2017,
  author    = {Lundberg, S. M. and Lee, S.-I.},
  title     = {A unified approach to interpreting model predictions},
  booktitle = {Advances in Neural Information Processing Systems},
  year      = {2017},
  volume    = {30}
}

@article{Ma2025,
  author  = {Ma, Jiacheng and Liu, Shengrui and Pei, Yuting},
  title   = {SHAP-based interpretable machine learning for injury risk prediction in university football players: a multi-dimensional data analysis approach},
  journal = {Scientific Reports},
  year    = {2025},
  volume  = {15},
  number  = {1},
  pages   = {40252},
  doi     = {10.1038/s41598-025-24144-y}
}

@article{MackenzieCushion2013,
  author  = {Mackenzie, R. and Cushion, C.},
  title   = {Performance analysis in football: A critical review and implications for future research},
  journal = {Journal of Sports Sciences},
  year    = {2013},
  volume  = {31},
  number  = {6},
  pages   = {639--676}
}

@book{Molnar2022,
  author    = {Molnar, C.},
  title     = {Interpretable Machine Learning: A Guide for Making Black Box Models Explainable},
  edition   = {2},
  publisher = {Leanpub},
  year      = {2022}
}

@article{Moustakidis2023,
  author  = {Moustakidis, Serafeim and Plakias, Spyridon and Kokkotis, Christos and Tsatalas, Themistoklis and Tsaopoulos, Dimitrios},
  title   = {Predicting football team performance with explainable AI: Leveraging SHAP to identify key team-level performance metrics},
  journal = {Future Internet},
  year    = {2023},
  volume  = {15},
  number  = {5},
  pages   = {174},
  doi     = {10.3390/fi15050174}
}

@article{Murtagh1991,
  author  = {Murtagh, F.},
  title   = {Multilayer perceptrons for classification and regression},
  journal = {Neurocomputing},
  year    = {1991},
  volume  = {2},
  number  = {5--6},
  pages   = {183--197}
}

@book{ODonoghue2009,
  author    = {O'Donoghue, P.},
  title     = {Research methods for sports performance analysis},
  publisher = {Routledge},
  year      = {2009}
}

@inproceedings{PantzalisTjortjis2020,
  author    = {Pantzalis, V. C. and Tjortjis, C.},
  title     = {Sports analytics for football league table and player performance prediction},
  booktitle = {2020 11th International Conference on Information, Intelligence, Systems and Applications (IISA)},
  year      = {2020},
  pages     = {1--8},
  publisher = {IEEE}
}

@article{Pappalardo2019,
  author    = {Pappalardo, L. and Cintia, P. and Rossi, A. and Massucco, E. and Ferragina, P. and Pedreschi, D. and Giannotti, F.},
  title     = {PlayeRank: Data-driven performance evaluation and player ranking in soccer via a machine learning approach},
  journal   = {ACM Transactions on Intelligent Systems and Technology},
  year      = {2019},
  volume    = {10},
  number    = {5},
  articleno = {59},
  numpages  = {27},
  doi       = {10.1145/3343172}
}

@inproceedings{Procopiou2023,
  author    = {Procopiou, Andria and Piki, Andriani},
  title     = {The 12th Player: Explainable Artificial Intelligence (XAI) in Football: Conceptualisation, Applications, Challenges and Future Directions},
  booktitle = {Proceedings of the 11th International Conference on Sport Sciences Research and Technology Support (icSPORTS)},
  year      = {2023},
  pages     = {213--220},
  doi       = {10.5220/0012233800003587}
}

@article{ReepBenjamin1968,
  author  = {Reep, C. and Benjamin, B.},
  title   = {Skill and chance in association football},
  journal = {Journal of the Royal Statistical Society: Series A (General)},
  year    = {1968},
  volume  = {131},
  number  = {4},
  pages   = {581--585}
}

@article{ReinMemmert2016,
  author  = {Rein, R. and Memmert, D.},
  title   = {Big data and tactical analysis in elite soccer: Future challenges and opportunities},
  journal = {SpringerPlus},
  year    = {2016},
  volume  = {5},
  number  = {1},
  pages   = {1--13}
}

@article{Sarmento2014,
  author  = {Sarmento, H. and Marcelino, R. and Anguera, M. T. and Campani{\c{c}}o, J. and Matos, N. and Leit{\~a}o, J. C.},
  title   = {Match analysis in football: A systematic review},
  journal = {Journal of Sports Sciences},
  year    = {2014},
  volume  = {32},
  number  = {20},
  pages   = {1831--1843}
}

@article{Sarmento2018,
  author  = {Sarmento, H. and Clemente, F. M. and Ara{\'u}jo, D. and Davids, K. and McRobert, A. and Figueiredo, A.},
  title   = {What performance analysts need to know about research trends in association football (2012--2016)},
  journal = {Sports Medicine},
  year    = {2018},
  volume  = {48},
  pages   = {799--836}
}

@inproceedings{Shrikumar2017,
  author    = {Shrikumar, A. and Greenside, P. and Kundaje, A.},
  title     = {Learning important features through propagating activation differences},
  booktitle = {Proceedings of the 34th International Conference on Machine Learning},
  year      = {2017},
  pages     = {3145--3153}
}

@article{Stafylidis2024,
  author  = {Stafylidis, A. and Mandroukas, A. and Michailidis, Y. and Vardakis, L. and Metaxas, I. and Kyranoudis, A. E. and Metaxas, T. I.},
  title   = {Key performance indicators predictive of success in soccer: A comprehensive analysis of the Greek Soccer League},
  journal = {Journal of Functional Morphology and Kinesiology},
  year    = {2024},
  volume  = {9},
  number  = {2},
  pages   = {107}
}

@article{Moya2025,
  title={Machine learning applied to professional football: Performance improvement and results prediction},
  author={Moya, Diego and Tipantu{\~n}a, Christian and Villa, G{\'e}nesis and Calder{\'o}n-Hinojosa, Xavier and Rivadeneira, Bel{\'e}n and {\'A}lvarez, Robin},
  journal={Machine Learning and Knowledge Extraction},
  volume={7},
  number={3},
  pages={85},
  year={2025},
  publisher={MDPI}
}

@article{Wang2024,
  author  = {Wang, Zhe and Velickovic, Petar and Hennes, Daniel and others},
  title   = {TacticAI: an AI assistant for football tactics},
  journal = {Nature Communications},
  year    = {2024},
  volume  = {15},
  pages   = {1906},
  doi     = {10.1038/s41467-024-45965-x}
}

@article{Yeung2025,
  author  = {Yeung, C. and Sit, T. and Fujii, K.},
  title   = {Transformer-based neural marked spatio-temporal point process model for analyzing football match events},
  journal = {Applied Intelligence},
  year    = {2025},
  volume  = {55},
  number  = {5},
  pages   = {4520--4536}
}

@article{Zhao2025,
  author  = {Zhao, Tingting and Cabral, Jeffrey and Zhu, Guangyu},
  title   = {A novel explainable artificial intelligence framework using knockoffs techniques with applications to sports analytics},
  journal = {Annals of Operations Research},
  year    = {2025},
  doi     = {10.1007/s10479-025-06575-y}
}

@article{Wong2025,
  title={A predictive analytics framework for forecasting soccer match outcomes using machine learning models},
  author={Wong, Albert and Li, Eugene and Le, Huan and Bhangu, Gurbir and Bhatia, Suveer},
  journal={Decision Analytics Journal},
  volume={14},
  pages={100537},
  year={2025},
  publisher={Elsevier}
}

@article{Hewitt2023,
  title={A machine learning approach for player and position adjusted expected goals in football (soccer)},
  author={Hewitt, James H and Karaku{\c{s}}, Oktay},
  journal={Franklin Open},
  volume={4},
  pages={100034},
  year={2023},
  publisher={Elsevier}
}

\end{document}